# Automated Sentiment and Hate Speech Analysis of Facebook Data by Employing Multilingual Transformer Models

R. Manuvie [1], S. Chatterjee [2]

[1] University College Groningen, University of Groningen, r.manuvie@rug.nl
[2] Foundation The London Story, Fluwelen Burgwal 58, 2511CJ, Den Haag, The Netherlands, saikat@thelondonstory.org

## Abstract

In recent years, there has been a heightened consensus – both within academia and in the public discourse – that Social Media Platforms (SMPs), amplify the spread of hateful and negative sentiment content. Researchers have identified how hateful content, political propaganda, and targeted messaging contributed to real-world harms including insurrections against democratically elected governments, genocide, and breakdown of social cohesion due to heightened negative discourse towards certain communities in parts of the world. To counter these issues, SMPs have created semi-automated systems that can help identify toxic speech. In this paper we analyse the statistical distribution of hateful and negative sentiment contents within a representative Facebook dataset (n= 604,703) scrapped through 648 public Facebook pages which identify themselves as proponents (and followers) of far-right Hindutva actors. These pages were identified manually using keyword searches on Facebook and on CrowdTangleand classified as far-right Hindutva pages based on page names, page descriptions, and discourses shared on these pages. We employ state-of-the-art, open-source XLM-T multilingual transformer-based language models to perform sentiment and hate speech analysis of the textual contents shared on these pages over a period of 5.5 years. The result shows the statistical distributions of the predicted sentiment and the hate speech labels; top actors, and top page categories. We further discuss the benchmark performances and limitations of these pre-trained language models.

## 1. Introduction

The spread of textual content with negative sentiment and community-targeted hate speech over Social Media Platforms (SMPs) like Facebook, Twitter, Reddit etc. has become a matter of serious concern in recent years (see Belew and Massanari 2018, Matamoros-Fernández and Farkas 2021, Bennett and Segerberg 2022). Case studies demonstrate a range of analytical results on the organic growth of hateful content around certain sensitive topics like a state election, COVID-19, the immigration policy of a country etc.; as well as amplified growth through coordinated rapid link-sharing behaviour by actor networks. In this study, our goal is to analyse the contents of a selected group of public Facebook fan pages (648 pages) which returned a total of 604,703 text messages on Facebook. The pages were manually identified as belonging to actors who are responsible for spreading anti-minority and anti-muslim narratives in contemporary India. An automated analysis of the textual contents associated with these fan pages was conducted to test the performance of current state-of-the-art NLP models in predicting and identifying contents with negative sentiment and hate speech respectively. Open-source multilingual Natural Language Processing (NLP) models for the sentiment analysis (Barbieri et al 2021), and the hate speech identification task (Röttger et al 2022) were employed in their 'evaluation' mode in order to perform a forward prediction task over the social media dataset (n=604,703). The approach allows a quantitative prediction of the amount of negative sentiment, and the amount of hateful content that is present in the dataset. Based on the outcome of these experiments, the statistical significance behind the abundance of hateful and negative sentiment content on the Facebook platform is discussed. Section 2 of the paper describes the process of data selection, preparation of the datasets, and our main research questions. Sections 3 and 4, discuss the results of sentiment and hate-speech analysis respectively. Section 5 is an extended discussion of our findings, concluding remarks and future scopes of this work.

## 2. Methodology and the dataset

The content analysis presented in this paper has been performed on Facebook's historical dataset that was ingested using the CrowdTangle platform in the date range of **31-12-2016** and **01-07-2022**. **648** Facebook pages were identified using keyword search as pages actively spreading anti-muslim and anti-minority hate speech in India (see Manuvie and Chaterjee, 2023 on the arXiv). The final dataset consists of a total of **604,703** entries, on which state-of-the-art open-source sentiment analysis and hate-speech detection models for forward predictions of the sentiment and hatefulness labels respectively were applied (see sections 3 and 4 below). The PyTorch code that exploits the transformers libraries and the Huggingface NLP models for these two forward inference tasks are publicly available at our GitHub repository.[1]

Based on the prediction scores of the NLP models, we investigate the following three main three research questions in this paper,

A. What are the distributions of sentiment and hate speech labels within our CrowdTangle dataset as predicted by the respective models?
B. Who are the top actors within our dataset who share hateful content and content with the negative sentiment?
C. What are the top categories of pages that are identified with hateful content and content with negative sentiment?

## 3. Sentiment analysis results

For sentiment analysis, the XLM-T multilingual sentiment analysis model from Cardiff NLP (Barbieri et al 2021) is employed.[2] The multilingual model is used as the Facebook textual dataset is a mixture of Hindi and English language texts. As XLM-T is a fine-tuned model of the XLM-roBERTa-base model, which is trained on ~198M tweets and further fine-tuned for sentiment analysis. According to the authors' benchmarking of the XLM-T multilingual model's performance on the Hindi language "Unified Multilingual Sentiment Analysis Benchmark" (UMSAB) dataset, the model has an F1 score of 56.39 (see Table 4 in Barbieri et al 2021). Although the performance of this model in the Hindi language is comparatively poor with respect to other languages (e.g., German and English have F1 scores of 77.35 and 70.63 respectively), the XLM-T model outperforms the XLM-R model by 7.9 absolute points in sentiment analysis task. Moreover, particularly because the model is trained on a corpus of social media datasets (i.e., tweets), we think this is relatively the best state-of-the-art open-source language model for our task of analyzing the sentiment of the mixed-language CrowdTangle dataset. From the results of the forward prediction, we see a distribution of **36.74%** negative, **40.07%** neutral and **23.19%** positive sentiment content respectively within our Facebook dataset which we have scrapped from CrowdTangle. The histogram in figure 1 demonstrates the percentages of these predicted labels.

---

[1] See the following GitHub page for the Google Colab notebooks on the prediction tasks: https://github.com/SaikatPhys/CrowdTangle-NLP.
[2] See the model card and the full model available at the Huggingface repository: https://huggingface.co/cardiffnlp/twitter-xlm-roberta-base-sentiment. Also check out their GitHub repository: https://github.com/cardiffnlp/xlm-t and the paper https://arxiv.org/pdf/2104.12250v2.pdf.

Figure 1: Distribution of sentiment analysis labels as predicted by CardiffNLP's XLM-T multilingual sentiment analysis model. The model predicts 37% negative, 40% neutral and 23% positive sentiment content within our CrowdTangle database.

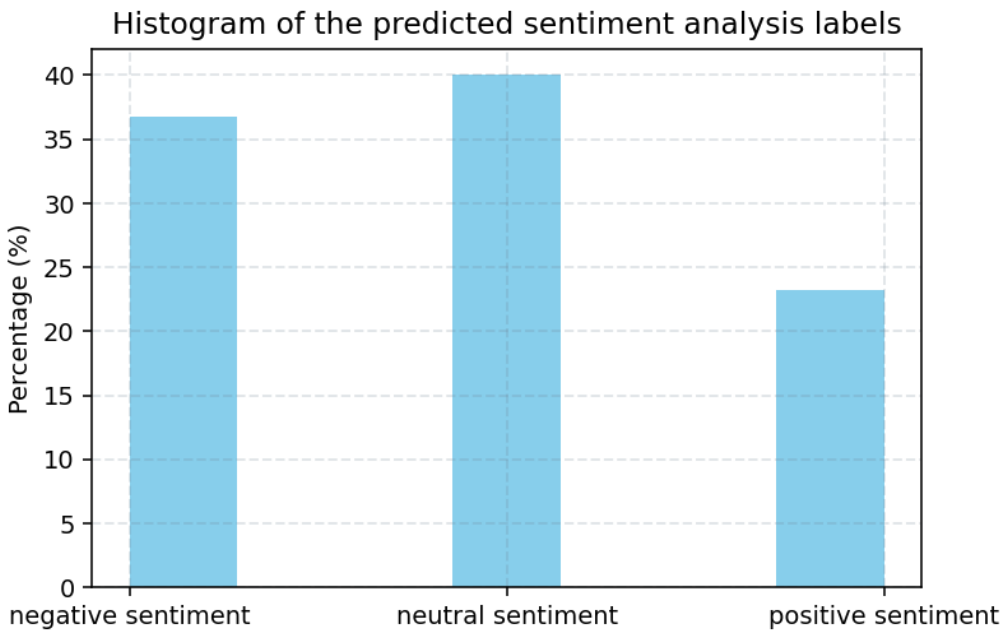

In order to identify the most dominant actors that are responsible for creating the negative sentiment content, the dataset of predicted 36.74% labels is split into the top 10 accounts (see figure 2). The top five Facebook pages within the negative sentiment set are found to be *"We Support Hindutva", "Pushpendra Kulshrestha Fans Club", "I Am Proud To Be A Hindu", "We support hindutva"* and *"Sanatan Press"* respectively. It's noteworthy that several of these pages have identical names however, this does not automatically imply identical ownership or identical content sharing behaviour. The top negative sentiment exhibiting pages are self-declared promoter/supporters of Hindutva (a far-right ideology which seeks to establish a Hindu nation in India) or supporters of Far-right political and ideological leaders. This is also because our database is collected from a subset of 648 far-right pages that were first manually selected. Within this limitation, we found the page *Sanatan Press* as one of the top 10 pages exhibiting negative sentiments. It is noteworthy that *Sanatan Press* is the media wing of *Sanatan Sanstha* a far-right organisation which has been identified as a dangerous organisation by Facebook and therefore banned on their platforms (TIME 2021).

Figure 3 plots the top 10 page categories which share the negative sentiment content. Amongst the 648 pages that were put through the scanner in this study, several of the pages self-identified as general activity pages, meaning that Facebook's content moderation policy should fully apply to these pages. Few pages identified as political organisations, politicians, Media News company and News sites. It is noteworthy that for political actors and news media, Facebook applies content moderation exceptions, meaning thereby that content shared by these pages is not as rigorously held to the content moderation policy. This has implications for automated and semi-automated content removal processes that we separately discuss elsewhere.

Figure 2. Predicted 36.74% of negative sentiment content is split into the top 10 actors in the descending order of their percentage of counts.

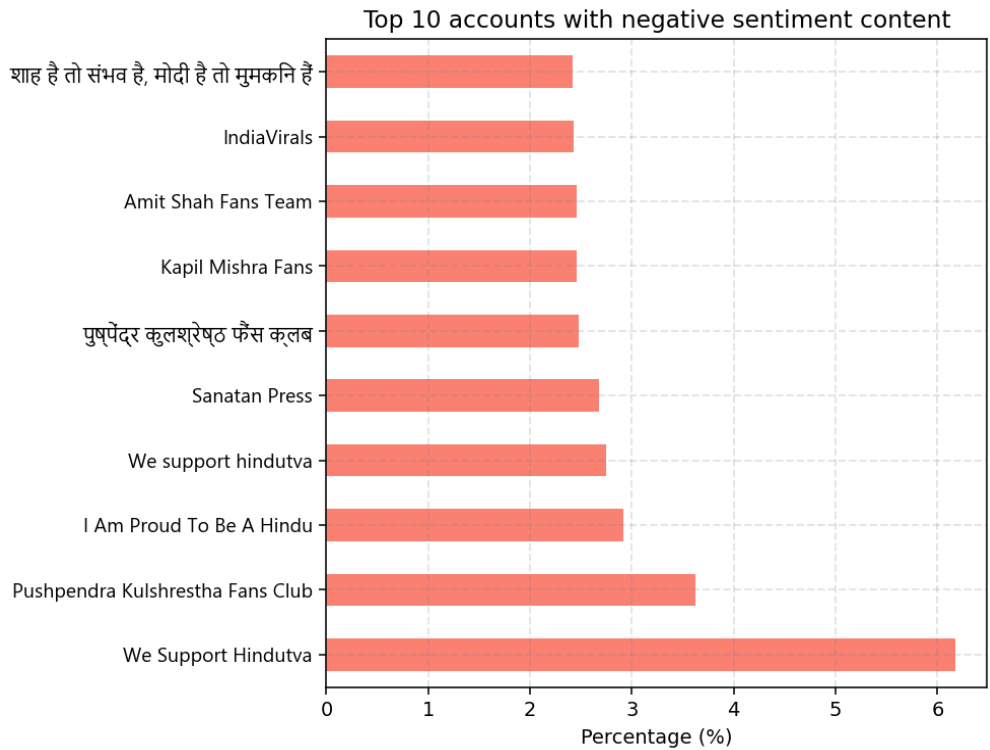

Figure 3: Top 10 page categories which share 36.74% of negative sentiment content.

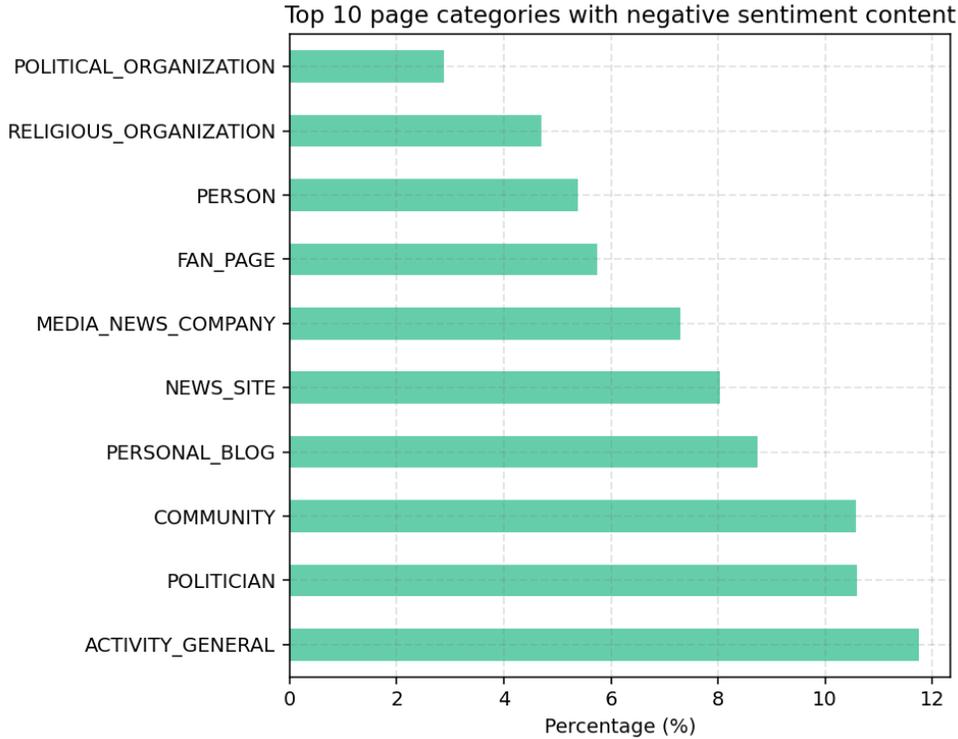

## 4. Hate speech detection results

To perform the hate speech detection task, the multilingual XTC model – a fine-tuned version of the XLM-T model on a multilingual hate speech dataset, was employed (Röttger et al 2022). The XTC model has an F1-score of 78.5 for the "hateful" label and an F1-score of 37.7 for the non-hateful label in the Hindi language benchmark dataset (see Table 2 in Röttger et al 2022). These scores imply that although the XTC model performs better in predicting hateful labels with fewer false positives, it can predict a significant amount of false negatives. Alternatively stated, it can predict hateful texts to be non-hateful.

In the current forward inference task, the XTC model predicts **11.71%** of the text contents to be hateful and **88.29%** to be non-hateful. Given the limitations of the F1 scores of the XTC model, we suspect that "non-hateful" content must be lesser in percentage than 88.29%. Because of the lower F1-score in the non-hateful label, it's probable that the XTC model has failed to rightly classify them as a "hateful" class. However, a higher F1 score in the hateful class allows us to confidently claim that almost all of this 11.71% hateful content is correctly predicted as hateful.

Figure 4. The distribution of predicted labels from the hate speech analysis model. The multilingual model predicts ~88% non-hateful and ~12% hateful content within the CrowdTangle dataset respectively

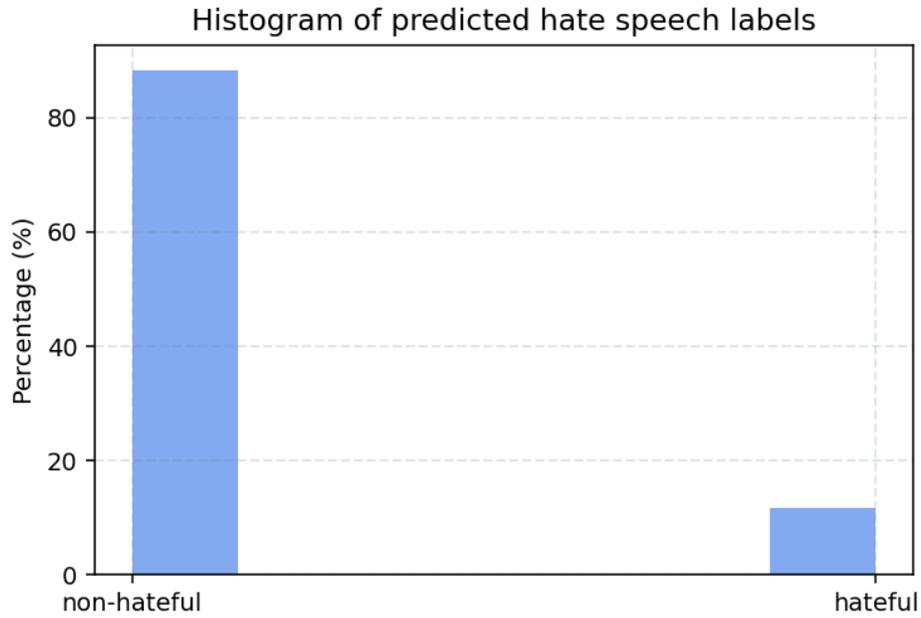

Figure 5. The predicted 11.71% of hateful content is further split into the top 10 actors in descending order.

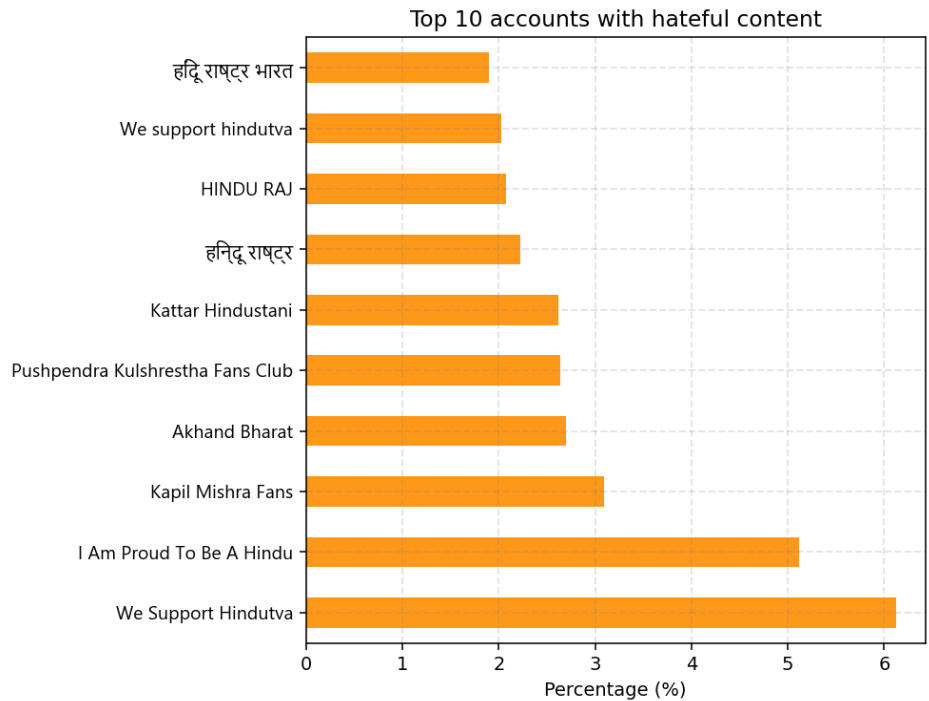

Figure 6. Top 10 page categories that share the 12% hateful content.

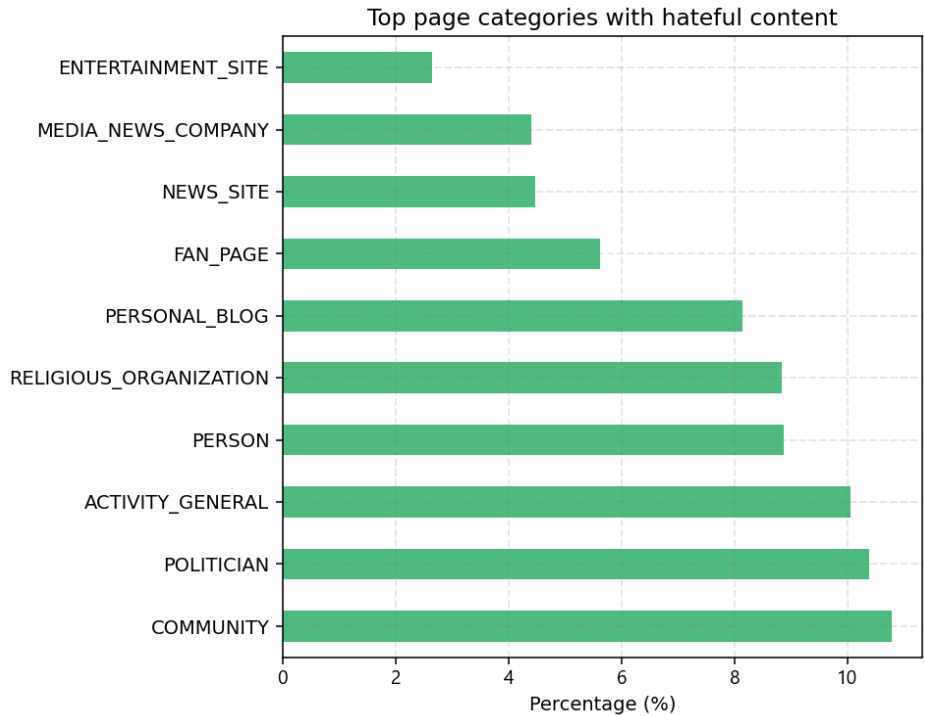

Table 1. Percentages of predicted labels in different categories of sentiment and hate (percentages are calculated w.r.t. the total size of the dataset, *N*=375,922)

| Labels | **Negative sentiment** | **Neutral sentiment** | **Positive sentiment** | Total |
|---|---|---|---|---|
| **Non-hateful** | 28.12% | 38.07% | 22.10% | 88.29% |
| **Hateful** | 8.62% | 2.00% | 1.09% | 11.71% |
| Total | 36.74% | 40.07% | 23.19% | 100% |

Given that the F1 score for the hateful label is 78.5, we have furthermore inspected the top accounts within our dataset that have shared the most hateful content (figure 5). The top five actors within the hateful content set are, *"We Support Hindutva", "I Am Proud To Be A Hindu", "Kapil Mishra Fans", "Akhand Bharat"* and *"Pushpendra Kulshrestha Fans Club"*. Three out of the top five actors (namely, *"We Support Hindutva", "I Am Proud To Be A Hindu"* and *"Pushpendra Kulshrestha Fans Club"*) are the same both in the sets of negative sentiment content and hateful datasets (see figures 2 and 5). As we further see in Table 1, out of the entire set of 11.71% identified hateful content, a majority of 8.62% belongs to the category of negative sentiment. However, the majority of negative sentiment content (i.e., 28.12% out of 36.74%) belongs to the non-hateful category. These statistics imply that negative sentiment content doesn't necessarily qualify to be hateful content but hateful content most likely qualifies to carry a negative sentiment. However, a match between the list of top actors in both the categories of hateful content and negative sentiment implies that most active perpetrators of negative sentiment content are also spreaders of hate speech.

In order to investigate what types of page categories are mostly associated with hateful texts, we have further split the hateful content into top page categories in Figure 6. If we compare the top 10 page categories between hate content and negative sentiment content, we find that including the ACTIVITY_GENERAL category, we have a match of altogether 9 categories (namely, COMMUNITY, FAN_PAGE, PERSONAL_BLOG, PERSON, POLITICIAN, NEWS_SITE, MEDIA_NEWS_COMPANY and RELIGIOUS_ORGANIZATION). This list shows what type of actors and what kind of pages act as the main perpetrators of the hateful and negative sentiment content within our dataset.

Table 2. Sample sizes of human annotations for validating the predicted labels of the hate speech model with a confidence level of 95% and a 5% margin of error.

| Labels | **Non-hateful** (Population size = 331,892) | **Hateful** (Population size = 44,030) | Total |
|---|---|---|---|
| **Sample size** | 384 | 381 | 765 |

In order to validate the accuracy of the hate speech model by human annotators, we used a set of randomly sampled 765 messages from the CrowdTangle dataset and annotated ourselves (see Table 2). In order for the annotations to be representative of the dataset, we used the number of model-predicted labels as our population sizes and kept a benchmark confidence level of 95% with a 5% margin of error. By employing these annotated labels as ground truth, we evaluated the XTC hate speech model's performance. During the process of annotation a non-binary scheme was adopted while allowed the annotators to label the text as hateful, non-hateful, insufficient context, abusive and negative discourse. The annotator had the ability to apply multiple labels to the text. Of the 765 datapoints which were re-labled the annotators marked 14% of messages as having "insufficient context"; and 6% as "abusive"; 23% of messages were identified as having "negative discourse". The rest of the dataset (~57%) was annotated as simply hateful or non-hateful. For the confusion matrix and evaluation report we are only using the data which marked as either hateful or non-hateful , as the model is unable to identify abusive, or discursive context only the binary labeling is used to re-train the model for the purpose of this paper.

The confusion matrix from the evaluation is shown in figure 7. The evaluation report of the model is shown in Table 3. The model shows F1-scores of 0.83 and 0.71 in the non-hateful and hateful labels respectively, which implies that the model is performing better in the hate speech prediction task after fine-tuning it through human annotation. Compared to the benchmark Hindi dataset used in the original paper - which had F1-scores of 0.78 and 0.37 respectively for these two labels the new F1 score is now 0.83 and 0.71 respectively. This already shows that the model is performing slightly better upon fine tunning, although inherently it still lacks nuances and contextual knowledge.

Figure 7. Normalised confusion matrix showing the performance of the hate speech model on annotated evaluation dataset

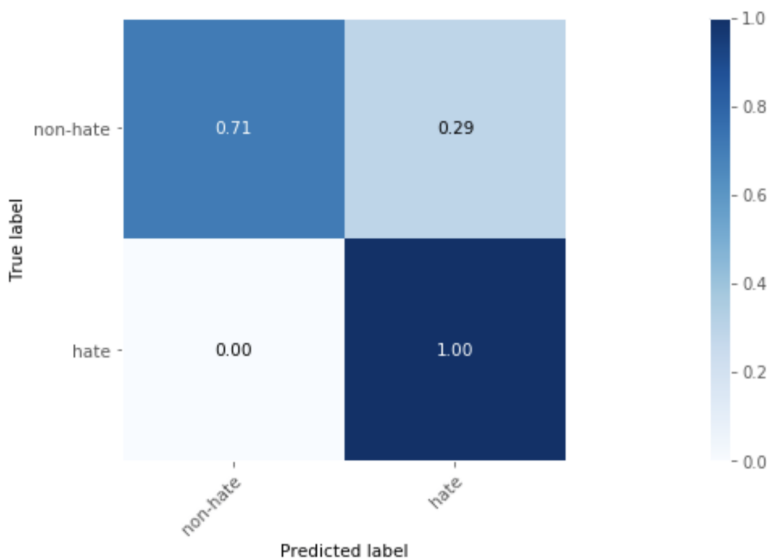

Table 3. Evaluation report showing the performance of the hate speech model

| Labels | Precision | Recall | F1-score |
| --- | --- | --- | --- |
| Non-hateful | 1.00 | 0.71 | 0.83 |
| Hateful | 0.55 | 1.00 | 0.71 |

## 5. Discussion and conclusions

While there is an evident correlation between hateful content and negative sentiment, from Table 1 we see that the majority of the non-hateful content is actually neutral in sentiment (i.e., 38.07% of 88.29%). Within the non-hateful category, the amount of negative sentiment content dominates over the amount of positive sentiment content (i.e, 28.12% compared to 22.10%) – which seems counter-intuitive. Any evident correlation between positive sentiment and non-hateful content from such statistical distributions of predicted labels can not be drawn. Given that the XTC model has an F1-score of 37.7 for the non-hateful label (as we mentioned earlier in section 4), it is suspected that several hateful texts are wrongly classified as non-hateful by the model. However, to what extent these false negatives are distributed amongst the three sentiment categories, is not quantifiable within this paper.

Factors that may contribute to the underperformance of the NLP models include definitions of hate speech and the categorization of negative/neutral/positive sentiment labels itself – which are implicitly encapsulated in the weights of the language models through the labelled datasets that were used for training. Our findings suggest that the state of art open-source technologies in sentiment analysis and hate speech analysis have an inherent limitation in the forward identification and detection of hate speech and negative sentiments. This has implications for the automated removal of content from social media platforms. While it is not known to the authors if similar predicting models are used by Very Large Social Media Platforms like Facebook, Twitter, and Youtube, it can be drawn that smaller and newer platforms which may rely on open-source technologies for sentiment analysis will encounter issues of false negative detection and may not be able to swiftly remove hateful content and negative sentiments from their platforms as effectively. This means that fully automated detection of hateful content especially in non-English and multilingual databases is far from being in a mature and deployable technology state.

As the above findings suggest, there is a need to further fine-tune the current state-of-the-art XLM-T class transformer models in popular downstream tasks like hate speech detection and sentiment analysis  – particularly in multilingual settings where the data consists of a mixed set of one Latin script (i.e., English) and one Devanagari script language (i.e., Hindi).

## Acknowledgements

The authors acknowledge the help received from the CrowdTangle – a Facebook-owned tool that tracks interactions on public content from Facebook pages and groups, verified profiles, Instagram accounts, and subreddits. The authors acknowledge the role of members of *Stichting The London Story* in constructive discussion on this paper.